%% file: root.tex
\documentclass[letterpaper, 10 pt, conference]{ieeeconf}  

\IEEEoverridecommandlockouts                              

\usepackage[pdftex]{graphicx}
\graphicspath{{figures/}}
\usepackage{hyperref}
\hypersetup{
	colorlinks=true,
	linkcolor=black,
	citecolor=black,
	filecolor=black,
	urlcolor=black,
}
\usepackage[T1]{fontenc}
\usepackage[utf8]{inputenc}
\usepackage{csquotes}
\usepackage[english]{babel}
\usepackage[export]{adjustbox}
\usepackage{caption}
\usepackage{subcaption}
\usepackage[colorinlistoftodos,backgroundcolor=orange!20,bordercolor=orange]{todonotes}
\usepackage{lipsum}
\usepackage[letterpaper,left=0.76in,right=.76in,top=0.72in,bottom=0.8in]{geometry}
\usepackage{amsmath}
\usepackage{amssymb}
\usepackage{mathtools}
\usepackage{calc}
\usepackage{pgfplots}
\usepackage{siunitx}

\usepackage{placeins}

\usepackage{float}
\usepackage{algorithm}
\usepackage{algorithmic}

\usepackage[style=ieee,
doi=false,
url=false,
mincitenames=1,
maxcitenames=1,
minbibnames=6,
maxbibnames=6,
backend=biber]{biblatex}  
\title{\LARGE \bf Large-Scale 3D Semantic Reconstruction for Automated Driving Vehicles with Adaptive Truncated Signed Distance
Function}

\author{
	Haohao Hu$^{\ast,1}$,
	Hexing Yang$^{\dagger,1}$,
	Jian Wu$^{\dagger,1}$,
	Xiao Lei$^{\dagger,1}$,
	Frank Bieder$^{2}$,
	Jan-Hendrik Pauls$^{1}$
	and Christoph Stiller$^{1, 2}$
\thanks{$^{\ast}$Corresponding author. $^{\dagger}$Authors contributed equally.}%
\thanks{$^{1}$ Authors are with the
\href{https://www.mrt.kit.edu/}{
	Institute of Measurement and Control Systems, 
	Karlsruhe Institute of Technology, 
	Karlsruhe, Germany. 
}
{\tt\small \{haohao.hu,jan-hendrik.pauls,stiller\}@kit.edu,
\{yanghexing93,leixiao\_94\}@outlook.com, 
\{xtdyzjc\}@163.com}}%
\thanks{$^{2}$ Authors are with the
\href{https://www.fzi.de/startseite/}{
	FZI Research Center for Information Technology,
	Karlsruhe, Germany. 
}
{\tt\small bieder@fzi.de}}%
}

\begin{document}

\maketitle
\thispagestyle{empty}
\pagestyle{empty}
\input{01_abstract}
\input{02_introduction}
\input{03_related_work}
\input{04_dense_mapping}
\input{05_texture_mapping}
\input{06_semantic_mapping}
\input{07_results_and_evaluation}
\input{08_conclusions}
\input{09_acknowledgements}
\printbibliography

\end{document}

%% file: 01_abstract.tex
\begin{abstract}
The Large-scale 3D reconstruction, texturing and semantic mapping are nowadays widely used for automated driving vehicles, virtual reality and automatic data generation.
However, most approaches are developed for RGB-D cameras with colored dense point clouds and not suitable for large-scale outdoor environments using sparse LiDAR point clouds.
Since a 3D surface can be usually observed from multiple camera images with different view poses, an optimal image patch selection for the texturing and an optimal semantic class estimation for the semantic mapping are still challenging.
To address these problems, we propose a novel 3D reconstruction, texturing and semantic mapping system using LiDAR and camera sensors.
An Adaptive Truncated Signed Distance Function is introduced to describe surfaces implicitly, which can deal with different LiDAR point sparsities and improve model quality.
The from this implicit function extracted triangle mesh map is then textured from a series of registered camera images by applying an optimal image patch selection strategy.
Besides that, a Markov Random Field-based data fusion approach is proposed to estimate the optimal semantic class for each triangle mesh.
Our approach is evaluated on a synthetic dataset, the KITTI dataset and a dataset recorded with our experimental vehicle.
The results show that the 3D models generated using our approach are more accurate in comparison to using other state-of-the-art approaches.
The texturing and semantic mapping achieve also very promising results.
\end{abstract}

%% file: 02_introduction.tex
\section{INTRODUCTION}
\label{sec:INTRODUCTION}
In recent years, the accurate 3D reconstruction, texturing and semantic mapping of surrounding environments has attracted more and more attention.
On the one hand, an accurately generated 3D geometric model of the real world extends functionalities of applications like virtual reality (VR) and augmented reality (AR).
On the other hand, for automated driving vehicles, an accurate 3D environment representation is essential for various tasks, such as mapping, localization, decision making and trajectory planning.
A realistic environment model also allows us to simulate different driving scenarios for the algorithm developing and testing in the field of autonomous driving, which can speed up the development sufficiently.
An accurately reconstructed and semantic mapped 3D model can accelerate the training data generation for deep learning without too much extra effort.
Besides that, the semantic mapped 3D models can also be used to extract semantic HD maps for automated driving vehicles automatically.
However, the large-scale 3D reconstruction, texturing and semantic mapping based on LiDAR and camera sensors remains challenging.
\begin{figure}
	\vspace{0.2cm}
	\centering
	\includegraphics[width = \columnwidth]{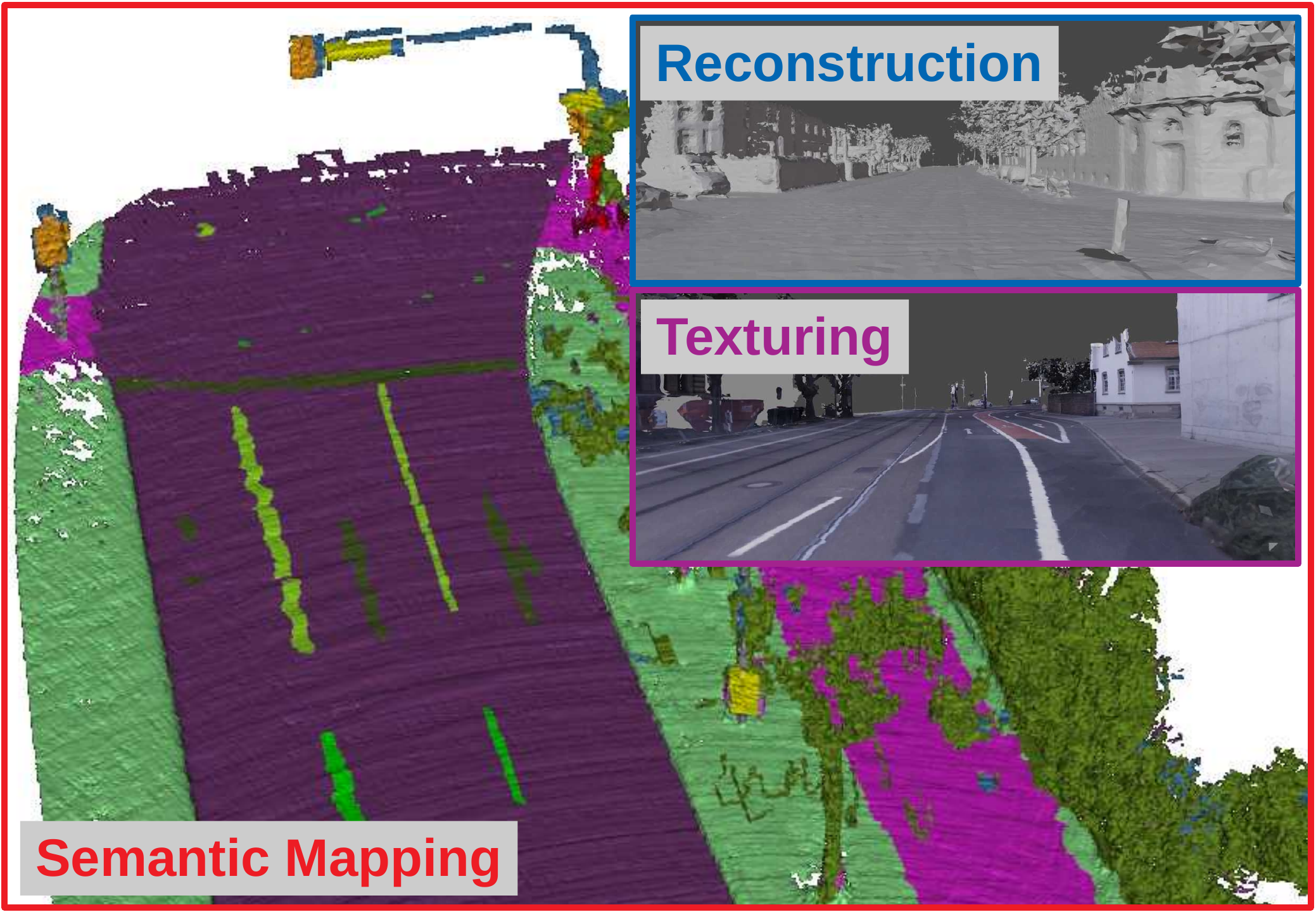}
	\caption{
	A result example of our approach including the 3D reconstructed triangle mesh map (in blue), the textured 3D model (in violet) and the semantic mapped 3D model (in red) based on LiDAR and camera measurements.
	}
	\label{fig:teaser}
\end{figure}
\begin{figure*}
	\vspace{0.2cm}
    \includegraphics[width=\textwidth]{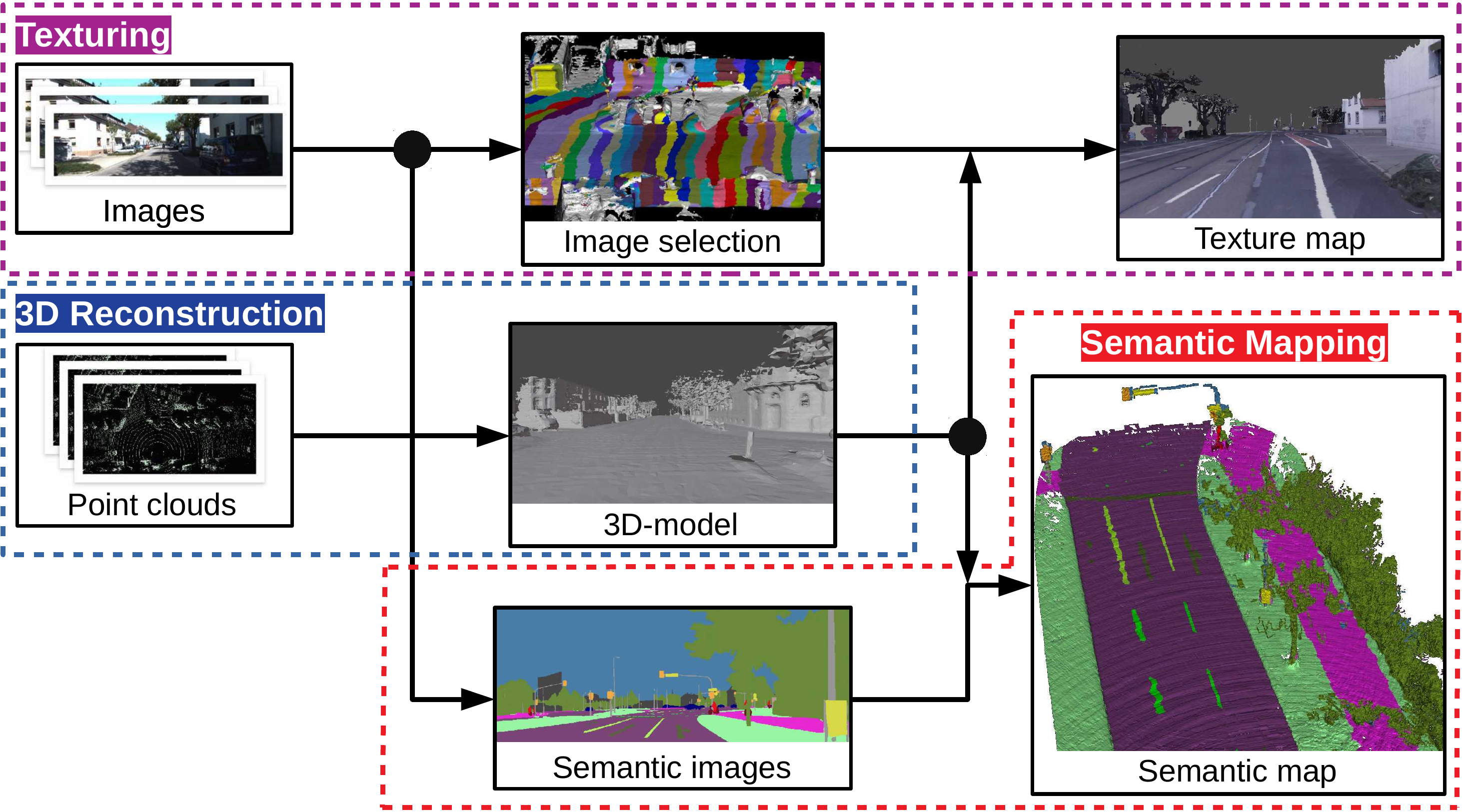}
    \caption{
	Our large-scale 3D reconstruction, texturing and semantic mapping pipeline for automated driving vehicles using LiDAR and camera sensors.
	An accurate 3D geometric environment model is obtained from accumulated LiDAR point clouds using Adaptive TSDF.
	The reconstructed 3D model is accurately textured from a series of registered camera images with an optimal image patch selection strategy and color optimization techniques.
	With a MRF-based data fusion approach, the 3D geometric model is then semantic mapped using multiple semantic segmented camera images.
    }
    \label{fig:flowchart}
\end{figure*}
In the past decade, research on dense 3D reconstruction has produced many compelling works using depth sensors based on the time-of-flight (ToF) or structured light technique \cite{izadi2011, whelan2012, niessner2013, Kahler2016, dai2017bundlefusion}.
Noteworthy, these for indoor scenes designed systems are not suitable for large-scale outdoor environments, since the measuring range of those depth sensors is small and the data quality is strongly influenced by outdoor light and weather conditions.
The recent advanced LiDARs enable another feasible solution for the 3D reconstruction of large-scale outdoor environments.
Compared with RGB-D cameras, the depth measurements of LiDARs are more accurate, reliable and reach a larger measuring range.
Nevertheless, the sparse and unevenly distributed LiDAR measurement points in the 3D space makes the 3D reconstruction very challenging.
Moreover, the reconstructed 3D geometric models from LiDAR point clouds do not contain any textures directly, which is however very useful.
Texturing geometric models from images is challenging because of the illumination and exposure time changes over time.
Besides that, an accurate semantic mapping of 3D geometric models is also challenging because of multiple observations of one triangle mesh.
In addition, moving objects in the input images make the texturing complex and uncertain.
The work presented in \cite{kuhner2020} performs a large-scale 3D reconstruction and texturing algorithm using LiDAR and camera sensors with Truncated Signed Distance Function (TSDF).
Our approach builds on this previous work and make several improvements.
Additionally, a semantic mapping pipeline of the reconstructed 3D models using semantic segmentation is also introduced, which is a perfect input for further applications like the automatic generation of semantic HD maps.
In this work, we propose a large-scale 3D reconstruction, texturing and semantic mapping approach based on LiDAR and camera measurements.
To handle the sparsity and heterogeneous density of the LiDAR point clouds, we propose an Adaptive Truncated Signed Distance Function (Adaptive TSDF)-based volumetric data fusion algorithm based on the well established work InfiniTAM \cite{Kahler2016}. 
Im comparison to the original Truncated Signed Distance Function (TSDF), our truncation distance is adaptively adjusted according to the density of LiDAR point measurements and the flatness of the reconstructed surfaces.
By benefiting from works \cite{lempitsky2007, waechter2014}, the reconstructed 3D geometric models can be textured with a very high resolution.
Through simplifying the optimization problem and parallelizing the part of the algorithm on a commercial GPU device, we can speed up the texturing process efficiently.
Secondly, applying an optimal image patch selection strategy and color optimization techniques improves the texturing quality of 3D models.
In addition, we use a Markov Ransom Filed (MRF)-based data fusion approach to realize an accurate semantic mapping of 3D models with multiple semantic segmented images.
We evaluate our approach on a synthetic dataset generated by LGSVL \cite{rong2020lgsvl}, the real world KITTI dataset \cite{Geiger2012CVPR} and also a real world dataset recorded with our experimental vehicle \textit{Bertha One} \cite{BerthaOne2018}.
Results show that our large-scale 3D reconstruction approach generates much smoother and more accurate 3D geometric models than our previous work.
The texturing and semantic mapping system provides also very accurate and promising results for further applications such as automatic semantic HD map generation.
The main contributions of this work are listed as follow:
\begin{enumerate}
	\item The Adaptive TSDF-based volumetric data fusion algorithm can deal with the heterogeneous density of LiDAR point clouds and improve the quality and continuousness of the reconstructed surfaces.
	\item The optimal image patch selection, consistency check and color optimization strategy improve the texturing quality and make our textured model seamless.
	\item The MRF-based data fusion approach improves the semantic class estimation for each reconstructed triangle face based on multiple semantic segmented images.
\end{enumerate} 

%% file: 03_related_work.tex
\section{RELATED WORK}
\label{sec:RELATED WORK}
\subsection{3D Geometric Reconstruction}
A suitable 3D environment representation of sensor measurements is a key to a high-quality 3D reconstruction.
In comparison to unstructured point-based \cite{deschaud2018,vlaminck2016,neuhaus2018} and surfel-based representations \cite{whelan2015, pfister2000}, volumetric methods with implicit surfaces have the most potential and are widely used on 3D reconstruction algorithms \cite{izadi2011, whelan2012, niessner2013, Kahler2016, dai2017bundlefusion}.
However, those approaches are developed for indoor environments using depth sensors like RGB-D cameras but not suitable for outdoor LiDAR scans because of the sparsity and heterogeneous density.
The work proposed in \cite{kuhner2020} extends KinectFusion \cite{izadi2011} to outdoor scenes using TSDF \cite{curless1996volumetric} with a large truncation distance at the cost of losing model details.
In \cite{roldao2019}, an adaptive neighborhood kernel based on Gaussian confidence evaluation is introduced.
Nevertheless, it's difficult to store statistics for all voxels due to memory limitation.

\subsection{Texture Mapping}
For indoor scenarios, frameworks based on RGB-D cameras \cite{izadi2011, Kahler2016, dai2017bundlefusion} obtain impressive textured models.
However, textures are encoded as either per-vertex color or per-face color and thus coupled to the mesh resolution.
For outdoor scenarios using LiDARs, the textured map can not be obtained directly.
The reconstructed 3D models are usually textured from a series of registered camera images.
Several approaches like \cite{callieri2008, grammatikopoulos2012} select and blend multiple images per face to obtain a global-consistent texture.
To solve this problem, the methods proposed in \cite{lempitsky2007, waechter2014} select only one image per face to texture 3D models.
If images are registered with small errors or the reconstructed models are slightly inaccurate, textures from different views might be misaligned, producing ghost and strong visible seams.
In this work, we apply an optimal image patch selection strategy to solve the single face multi-view image issue.
Besides that, several color optimization techniques are introduced to reduce the texture seams and improve the texture continuousness.

\subsection{Semantic Mapping}
The 3D semantic reconstruction are divided into indoor and outdoor scenes.
For indoor scenes, individual pre-defined object models \cite{salas2013slam++}, volumetric models \cite{mccormac2018fusion++} \cite{grinvald2019volumetric}, meshes \cite{feng2019localization} or quadrics \cite{nicholson2018quadricslam} are used.
However, outdoor scenes can not be represented by distinct components or simple geometric primitives.
Continuous structures like road markers, barriers, and buildings are represented in different ways. 
The authors in \cite{bordes2013information} assess the scene in terms of its primitives, such as ground, vegetation and barriers with the Dempster Shafer theory on a 2D image plane.
The approach presented in \cite{rosinol2020kimera} extends the work \cite{oleynikova2017voxblox} to a 3D semantic reconstruction, which deals with indoor environments and works with a limited sensing range, resolution and quality.
Instead of using 3D mesh maps, some approaches like \cite{herb2019crowd} \cite{herb2021lightweight} use outer contours of semantic edges to reconstruct 3D edges directly, which needs less resources.
However, the reduced mesh resolution also limits the reconstruction quality.

%% file: 04_dense_mapping.tex
\section{3D Geometric Reconstruction}
\label{sec:3D Geometric Reconstruction}
\begin{figure}
    \centering
    \vspace{0.2cm}
    \includegraphics[width = 0.65\columnwidth]{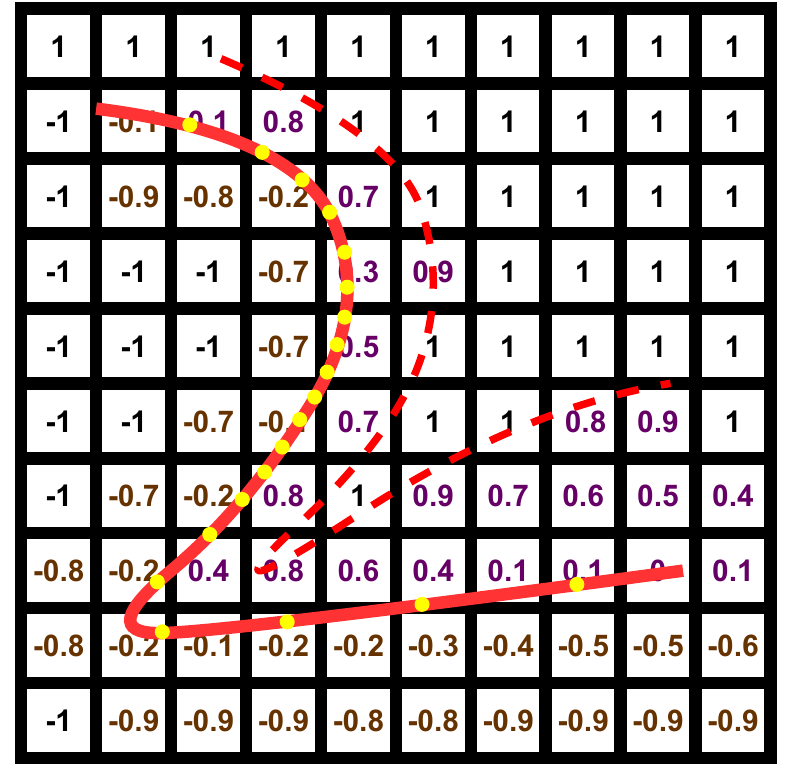}
    \caption{
        Adaptive TSDF-based implicit surface representation.
        Each yellow dot represents a measurement point, the red solid line corresponds to the reconstructed surface and the numbers in grids are TSDF values.
        To ensure the completeness of the reconstructed surface and retain as much detail as possible, the truncation distance $\varepsilon$ is dynamically adapted to the point measurements as shown in the red dotted line.
    }
    \label{fig:implicit_surface_representation}
\end{figure}
Similar as KinectFusion \cite{izadi2011}, we update the implicit surface representation of measurements at each frame parallel.
The voxel block hashing contributed by \cite{niessner2013, Kahler2016} allows for a volumetric fusion at scale.
The basic technique used here for the 3D reconstruction is TSDF and one of the most important parameter is called truncation distance $\varepsilon$, which denotes the influence range of a measurement point on the updating of the 3D voxel array.
By using a small $\varepsilon$, the completeness and continuousness of the reconstructed surfaces can be worsened because of a low point cloud density. 
A big $\varepsilon$ makes the reconstructed surfaces smoother but the details, where the structure is complex and narrowly, won't be reconstructed accurately.
Choosing a suitable $\varepsilon$ for different environments is always challenging, because of the various of the point cloud density and environment complexity.
To solve this problem, as illustrated in Fig. \ref{fig:implicit_surface_representation}, we introduce a dynamically computed truncation distance $\varepsilon$ based on the point cloud density and structure complexity adaptively, which allows to cope with the sparsity and heterogeneous density of LiDAR point clouds.
Most importantly, to tackle the heterogeneous density of the input data, we calculate the 3D statistical distribution of points for each voxel block consisting of $8\times8\times8$ voxels, rather than for each voxel as proposed in \cite{roldao2019}.
This concession allows us to keep a good trade-off between rich information and memory consumption.

\subsection{Planar Surface Estimation}
\label{sec:Planar Surface Estimation}
As we are limited by the physical memory, we calculate the statistical distribution upon insertion of new measurement incrementally.
We consider a point cloud consists of $n$ points $\mathcal{P}\!\to\!\{p_\mathrm{0}, p_\mathrm{1},...,p_\mathrm{n}\}$, the 3D statistical distribution includes the number of points $n$, the mean $E_\mathrm{\mu}$ and the covariance matrix $C_\mathrm{\sigma}$.
Suppose we have two point clouds $\mathcal{P}_\mathrm{1}\!\to\!\{p_\mathrm{0}, p_\mathrm{1},...,p_\mathrm{n_1}\}$, $\mathcal{P}_\mathrm{2}\!\to\!\{p_\mathrm{0}, p_\mathrm{1},...,p_\mathrm{n_2}\}$, where $n_\mathrm{1}$, $E^\mathrm{1}_\mathrm{\mu}$, $C^\mathrm{1}_\mathrm{\sigma}$ and $n_\mathrm{2}$, $E^\mathrm{2}_\mathrm{\mu}$, $C^\mathrm{2}_\mathrm{\sigma}$ represents its distribution information respectively.
For the sum of the two point clouds, the new distribution $E_\mathrm{\mu}'$, $C_\mathrm{\sigma}'$ can be computed by:
\begin{equation}
\begin{split}
    E_\mathrm{\mu}' & = \frac{n_\mathrm{1}E_\mathrm{\mu}^\mathrm{1} + n_\mathrm{2}E_\mathrm{\mu}^\mathrm{2}}{n_\mathrm{1} + n_\mathrm{2}} \\
    C_\mathrm{\sigma}' & = \frac{1}{n_\mathrm{1} + n_\mathrm{2} - 1}\{(n_\mathrm{1} - 1)C_\mathrm{\sigma}^\mathrm{1} + (n_\mathrm{2} - 1)C_\mathrm{\sigma}^\mathrm{2} \\
        & \quad + n_\mathrm{1}(E_\mathrm{\mu}' - E_\mathrm{\mu}^\mathrm{1})(E_\mathrm{\mu}' - E_\mathrm{\mu}^\mathrm{1})^\intercal \\
        & \quad + n_\mathrm{2}(E_\mathrm{\mu}' - E_\mathrm{\mu}^\mathrm{2})(E_\mathrm{\mu}' - E_\mathrm{\mu}^\mathrm{2})^\intercal\} \text{.}
\end{split}
\end{equation}
Especially for a point cloud just with exactly one point $\mathcal{P} \to \{p_\mathrm{0}\}$ hold $E_\mathrm{\mu} = p_\mathrm{0}$ and $C_\mathrm{\sigma} = [0]_\mathrm{3\times3}$.
After having obtained the local statistics, the covariance matrix $C_\mathrm{\sigma}$ can be used to estimate the local surface through a Principle Component Analysis (PCA).
Suppose the eigenvectors $(\vec{v}_\mathrm{1}, \vec{v}_\mathrm{2}, \vec{v}_\mathrm{3})$ and the eigenvalues $(\lambda_\mathrm{1}, \lambda_\mathrm{2}, \lambda_\mathrm{3})$ with $\lambda_\mathrm{1}>\lambda_\mathrm{2}>\lambda_\mathrm{3}$, the least dominant axis $\vec{v}_\mathrm{3}$ is used to represent the unoriented normal direction of the estimated surface.
Since the normal direction should be consistent and oriented towards the sensor, the normal of the planar surface can be written as:
\begin{equation}
\begin{split}
    \vec{n} = \left\{
         \begin{array}{lr}
         \vec{v}_\mathrm{3} \ \ \quad \text{if}\ \vec{v}_\mathrm{3}\cdot (\vec{p}_\mathrm{s} - E_\mathrm{\mu}) > 0 \\
        -\vec{v}_\mathrm{3} \quad \text{otherwise,}
         \end{array}
    \right.
\end{split}
\label{equ:normal_direction}
\end{equation}
where $\vec{p}_\mathrm{s}$ is the 3D position of the sensor.
Furthermore, the flatness of the surface $P_\mathrm{flat}$ can be represented as:
\begin{equation}
    P_\mathrm{flat} = 1-\frac{\lambda_\mathrm{3}}{\lambda_\mathrm{2}} \text{.}
\end{equation}

\subsection{Implicit Surface Representation}
\label{sec:Implicit Surface Representation}
As described before, large truncation distance $\varepsilon$ ensures the surface's completeness and a small $\varepsilon$ preserves much model details.
Therefore, we adaptively adjust the $\varepsilon$ according to the point cloud density and surface flatness using:
\begin{equation}
    \varepsilon = \max(\varepsilon_\mathrm{min}, \frac{kP_\mathrm{flat}}{n}\varepsilon_\mathrm{max}) \text{,}
\end{equation}
where $\varepsilon_\mathrm{min}$ and $\varepsilon_\mathrm{max}$ are the lower and upper bound, $n$ is the number of measurement points inside the voxel block and $k$ is the weight to describe the importance of the flatness.
Additionally, to reduce the effects of outliers and systematic distortions of range sensors, a weight $w$ for each measurement is calculated using $w = \vec{n} \cdot (\vec{p}_\mathrm{s} - \vec{p}_\mathrm{i})$, where $\vec{n}$ is the surface normal, $\vec{p}_\mathrm{s}$ is the sensor position and $\vec{p}_\mathrm{i}$ represents a measurement point.
The basic idea is that a lower angle between the incident light and the normal of the surface usually corresponds to better measurement samples.  

%% file: 05_texture_mapping.tex
\section{Texture Mapping}
\label{sec:Texture Mapping}
\subsection{Image Selection}
\label{sec:Image Selection}
To obtain texture maps with a high resolution, we texture each face with the best image patch rather than select and blend multiple images. 
To check the face visibility for all combinations of images and faces in parallel, we implement a BVH tree \cite{karras2012} in CUDA \cite{cuda2020}.
For each face, we select the best image patch using a MRF-based energy function \cite{lempitsky2007}:
\begin{equation}
\sum_{F_\mathrm{i} \in \text{faces}} E_\mathrm{Q}(F_\mathrm{i}, \mathbf{I}_\mathrm{i}) + \lambda \sum_{F_\mathrm{i},F_\mathrm{j} \in \text{edges}}E_\mathrm{S}(F_\mathrm{i},F_\mathrm{j},\mathbf{I}_\mathrm{i},\mathbf{I}_\mathrm{j}) \text{,}
\end{equation}
where $E_\mathrm{Q}$ represents the quality of the image $\mathbf{I}_\mathrm{i}$ to be used for face $F_\mathrm{i}$ and $E_\mathrm{s}$ corresponds to the smoothness between two adjacent faces textured from different images.
The work in \cite{lempitsky2007} uses the angle between the local viewing direction and the face normal as the cost.
However, that is insufficient for us since the vehicle motion makes the angle difference not obvious.
In \cite{allene2008, buehler2001}, a face is projected into image and the projection area size is used as the quality term or the blending weight.
Nevertheless, this does not account for extreme cases such as underexposure or overexposure of the recorded images.
The authors of \cite{waechter2014} use both the gradient magnitude of the image and the face's projection area size to handle out-of-focus blur.
We choose a similar term since it accounts for similar effects and thus the first term can be written as the sum of the gradient magnitude $||\Delta(\mathbf{I}_\mathrm{i}(p))||_\mathrm{2}$ of all pixels within the projection area:
\begin{equation}
    E_\mathrm{Q}(F_\mathrm{i},\mathbf{I}_\mathrm{i}) = \int_{\phi(F_\mathrm{i}, \mathbf{I}_\mathrm{i})}||\Delta(\mathbf{I}_\mathrm{i}(p))||_\mathrm{2} \mathrm{d}p \text{.}
\end{equation}
Considering two adjacent faces $F_\mathrm{i}$ and $F_\mathrm{j}$ share an edge $E_\mathrm{ij}$, the color seam between them can be evaluated by integrating all pixel differences along this edge \cite{lempitsky2007}.
However, this is a significant bottleneck of this algorithm and taking account of by selecting the same image for adjacent faces can achieve the same goal with the simplified term:
\begin{equation}
    E_\mathrm{S}(F_\mathrm{i}, F_\mathrm{j}, \mathbf{I}_\mathrm{i}, \mathbf{I}_\mathrm{j}) = \left\{
             \begin{array}{lr}
             1 \quad \text{if} \ \mathbf{I}_\mathrm{i} = \mathbf{I}_\mathrm{j}  \\
             0 \quad \text{otherwise.}
             \end{array}
    \right.
\end{equation}

\subsection{Color Adjustment}
\label{sec:Color Adjustment}
Due to color discontinuities or wrong textures caused by occlusions like moving objects, more complex processing steps are necessary to optimize the textures.

\subsubsection{Vignetting Correction}
The color discontinuities are mainly caused by the vignetting of images, where the brightness or saturation of pixels around the image drop significantly compared to the image center.
To reduce this effect, we use a vignetting correction algorithm \cite{rohlfing2012}:
\begin{equation}
    g(r) = 1+ar^2 + br^4 + cr^6
\end{equation}
where $g(r)$ is the gain and $r$ corresponds to the distance from the image center $(\Bar{u}, \Bar{v})$ to the pixel being treated $(u,v)$:
\begin{equation}
    r = \frac{\sqrt{(u-\Bar{u})^2 + (v-\Bar{v})^2}}{\sqrt{\Bar{u}^2 + \Bar{v}^2}}
\end{equation}
After choosing a set of suitable parameters $a$, $b$ and $c$, the adjusted image without vignetting can be obtained by:
\begin{equation}
    \mathbf{I}'(u, v) = g(r) \times \mathbf{I}(u,v)
\end{equation}

\subsubsection{Photo Consistency Check}
To remove color outliers, the photo consistency check algorithm proposed in \cite{waechter2014} is applied in the HSV color space.
This algorithm assumes that the majority of images have the correct color.

\subsubsection{Global Color Adjustment}
To make our textured maps as seamless as possible, we use a continuous formulation of global color adjustment to further improve the appearance of the texture using the approach proposed in \cite{lempitsky2007}.
To minimize color discontinuities, we compute an additive correction $g_\mathrm{v}$ as in \cite{lempitsky2007} for each vertex by minimizing:
\begin{equation}
    \sum_{v_\mathrm{i}}(f_{v_\mathrm{i}^\mathrm{l}} + g_{v_\mathrm{i}^\mathrm{l}} - (f_{v_\mathrm{i}^\mathrm{r}} + g_{v_\mathrm{i}^\mathrm{r}}))^2 + \lambda\sum_{v_\mathrm{i}, v_\mathrm{j} \in \text{edges}}(g_{v_\mathrm{i}^\mathrm{l}} - g_{v_\mathrm{j}^\mathrm{r}})^2
    \label{equ:seam_levelling}
\end{equation}
where $f_{v_\mathrm{i}^\mathrm{l}}$ is calculated by averaging color samples along the edge $E_{v_\mathrm{i}v_\mathrm{j}}$.
The first term in Eq. \ref{equ:seam_levelling} ensures that the colors on both sides of an edge are as similar as possible, while the second term aims at finding the smallest absolute additive correction.
After finding optimal $g_\mathrm{v}$ for all vertices, the correction for each pixel inside a face is interpolated using barycentric coordinates.

%% file: 06_semantic_mapping.tex
\section{Semantic Mapping}
\label{sec:semantic}

\begin{figure}
    \vspace{0.2cm}
    \centering
    \includegraphics[width=\columnwidth]{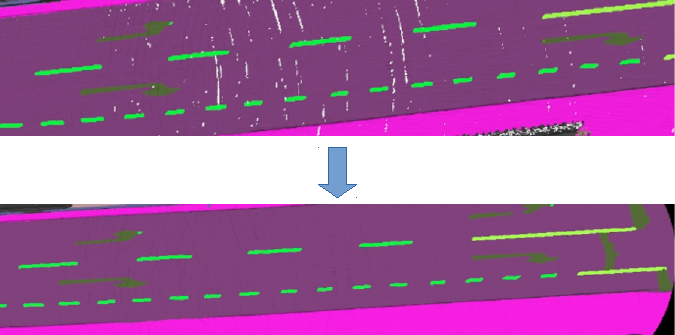}
    \caption{Elimination of invisible faces.}
    \label{fig:white}
\end{figure}
In addition to a textured geometric model, semantic information is also critical to interpret surrounding environments.
If the input source in the texturing procedure are changed to semantic segmented images, we can get a colorized mesh map, in which the color of each face represents a semantic class.
The only difference to the previous step presented in Sec. \ref{sec:Image Selection} is the label here corresponds to semantic classes.
The face visibility is checked with a BVH tree \cite{karras2012}.
We use a MRF-based energy function \cite{lempitsky2007} to select the face class:
\begin{equation}
\sum_{F_\mathrm{i} \in \mathrm{faces}} \!\! E_\mathrm{D}(F_\mathrm{i}, C_\mathrm{m}) + \lambda \!\! \sum_{F_\mathrm{i},F_\mathrm{j} \in \mathrm{edges}} \!\! E_\mathrm{S}(F_\mathrm{i},F_\mathrm{j},C_\mathrm{m},C_\mathrm{n}) \text{,}
\end{equation}
 where $E_\mathrm{D}$ denotes the data term and represents the probability of the face $F_\mathrm{i}$ to be class $C_\mathrm{m}$ concerning the semantic image data.
 $E_\mathrm{s}$ corresponds to the smoothness between two adjacent faces.
 To calculate the probability of each face, we project the face into image and use the size of the projected face as the data term like in \cite{allene2008}.
 Since a face is visible in multiple frames, the accumulated size after normalization is used as the final probability.
 To deal with faces on the edge, a part of which belongs to $C_\mathrm{m}$ and another part belongs to $C_\mathrm{n}$, we only consider the center pixel of a triangle.
 The area of the triangle is calculated as the number of votes:
 \begin{equation}
    E_\mathrm{D}(F_\mathrm{i},C_\mathrm{m}) = \frac{\sum\limits_{I_\mathrm{k} \in \mathrm{visible}} S_{I_\mathrm{k}}(F_\mathrm{i},C_\mathrm{m})}{\sum\limits_{m=0}^{N} \sum\limits_{I_\mathrm{k} \in \mathrm{visible}} S_{I_\mathrm{k}}(F_\mathrm{i},C_\mathrm{m})} \text{,}
    \label{equ:semantic_mrf_energy_function}
\end{equation}
where $N$ is the total number of classes and $I_\mathrm{k}$ is one of the visible frames after occlusion detection with the BVH tree. 
$S_{I_\mathrm{k}}(F_\mathrm{i},C_\mathrm{m})$ is the size of face $F_\mathrm{i}$ in frame $I_\mathrm{k}$, which is labelled as class $C_\mathrm{m}$.
Another issue is to deal with invisible faces as illustrated in Fig. \ref{fig:white}.
To address this problem, we adjust the smoothing term $E_\mathrm{s}$ in our MRF-based energy function Eq. \ref{equ:semantic_mrf_energy_function}.
The class probability of each face is initialized with an uniform distribution to all classes, which means $E_\mathrm{D}\left(C_\mathrm{1}\right)=E_\mathrm{D}\left(C_{2}\right)=\ldots=E_\mathrm{D}\left(C_\mathrm{N}\right)=\frac{1}{N}$.
Then let the smoothing term $E_\mathrm{S}$ judge, which class the face gets.

%% file: 07_results_and_evaluation.tex
\section{EXPERIMENTAL EVALUATION}
\label{sec:EXPERIMENTAL EVALUATION}
We evaluate our 3D reconstruction, texturing and semantic mapping system on: a synthetic dataset from LGSVL\cite{rong2020lgsvl}, the real world KITTI dataset \cite{Geiger2012CVPR} and a dataset recorded with our experimental vehicle \textit{Bertha One} \cite{BerthaOne2018} equipped with a Velodyne VLS-128 LiDAR and cameras.

\subsection{3D Geometric Reconstruction}
\begin{figure}
    \vspace{0.2cm}
    \begin{subfigure}{0.32\columnwidth}
    \includegraphics[width=\columnwidth]{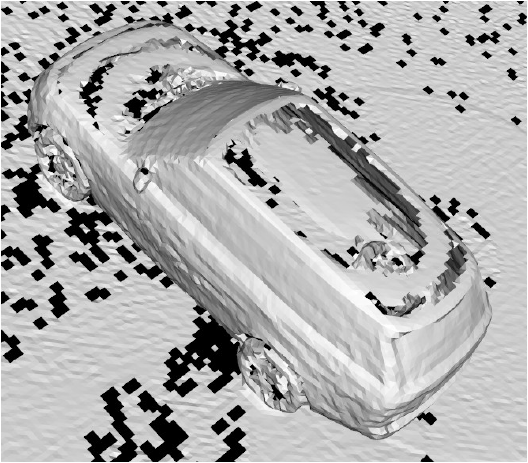}
    \subcaption{$\varepsilon = 10$ cm}
    \label{fig:suv_a}
    \end{subfigure}
    \begin{subfigure}{0.32\columnwidth}
    \includegraphics[width=\columnwidth]{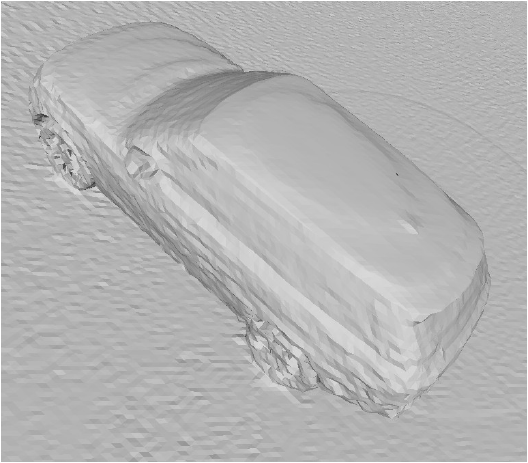}
    \subcaption{$\varepsilon = 30$ cm}
    \label{fig:suv_b}
    \end{subfigure}
    \begin{subfigure}{0.32\columnwidth}
    \includegraphics[width=\columnwidth]{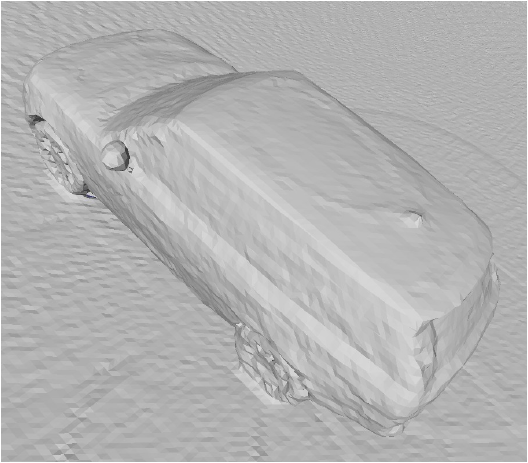}
    \subcaption{$\varepsilon_\mathrm{apt}:[10, 30]$ cm}
    \label{fig:suv_c}
    \end{subfigure}
    \caption{
    3D model of a SUV with a voxel size of $2.5$ cm.
    }
    \label{fig:reconstruction_suv}
\end{figure}

\begin{figure}
    \vspace{0.2cm}
    \centering
    \begin{subfigure}{\columnwidth}
    \includegraphics[width=\columnwidth]{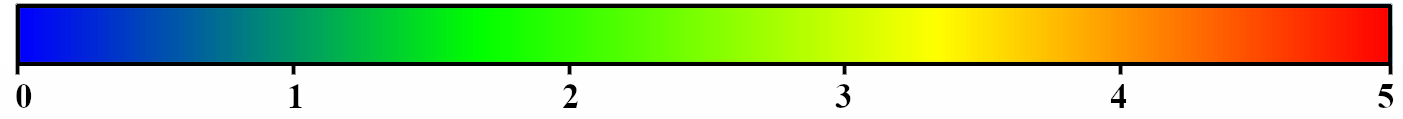}
    \subcaption*{Error [cm]}
    \end{subfigure}
    \begin{subfigure}{\columnwidth}
    \includegraphics[width=0.49\columnwidth]{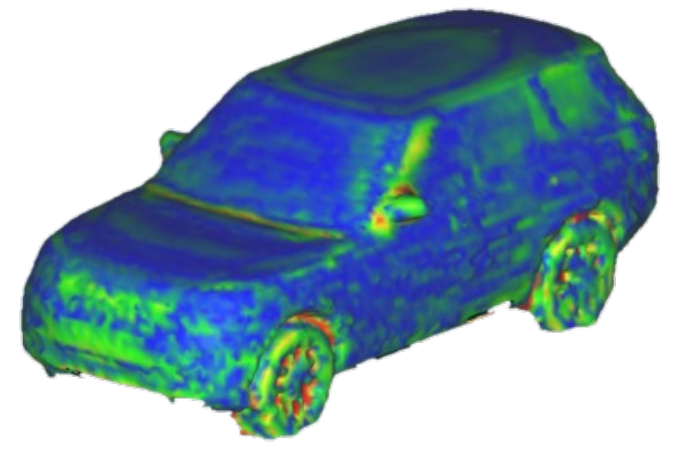}
    \includegraphics[width=0.49\columnwidth]{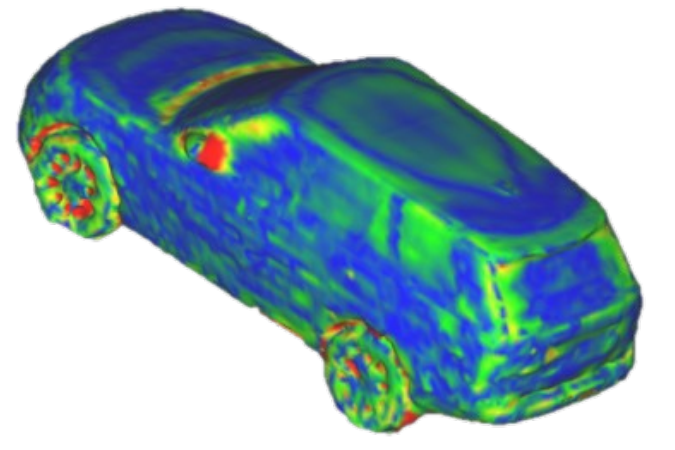}
    \subcaption{$\varepsilon = 30$ cm}
    \label{fig:suv_a_comp}
    \end{subfigure}
    \begin{subfigure}{\columnwidth}
    \includegraphics[width=0.49\columnwidth]{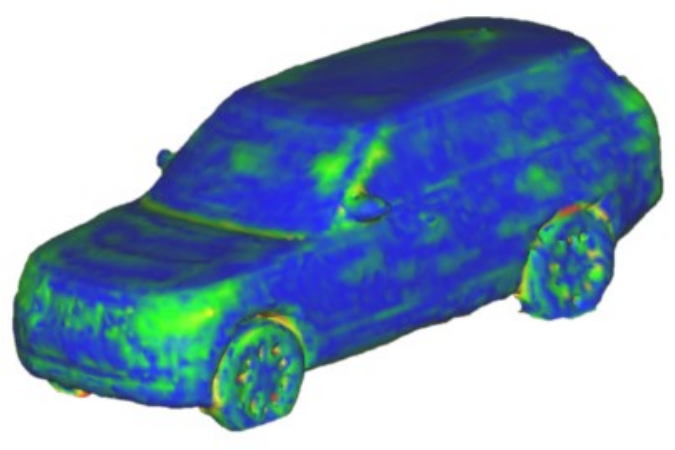}
    \includegraphics[width=0.49\columnwidth]{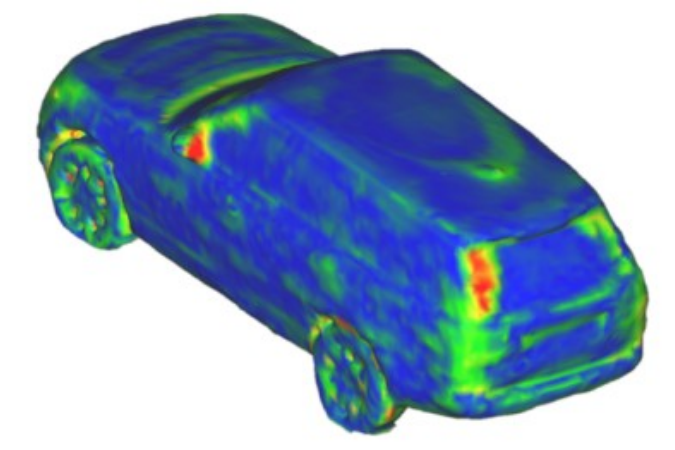}
    \subcaption{$\varepsilon_\mathrm{apt}:[10, 30]$ cm}
    \label{fig:suv_b_comp}
    \end{subfigure}
    \caption{
    Accuracy evaluation of reconstructed models.
    }
    \label{fig:reconstruction_suv_comp}
\end{figure}

\begin{figure}
    \vspace{0.2cm}
    \centering
    \begin{subfigure}{\columnwidth}
    \includegraphics[width=\columnwidth]{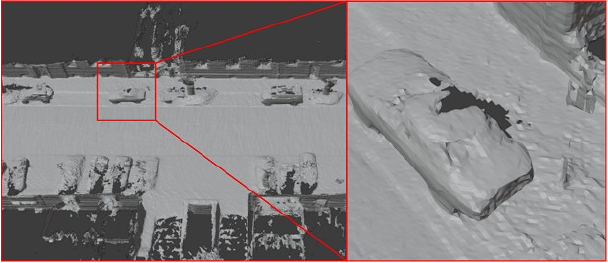}
    \subcaption{$\varepsilon=80$ cm}
    \label{fig:kitti_tsdf}
    \end{subfigure}
    \begin{subfigure}{\columnwidth}
    \includegraphics[width=\columnwidth]{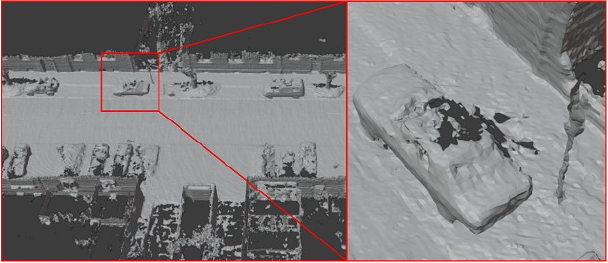}
    \subcaption{$\varepsilon_\mathrm{apt}:[40, 80]$ cm}
    \label{fig:kitti_atsdf}
    \end{subfigure}
    \caption{
    3D reconstruction on the KITTI dataset with a voxel size of $10$ cm and different truncation distance $\varepsilon$.
    }
    \label{fig:reconstruction_kitti}
\end{figure}

\begin{figure}
    \centering
    \begin{subfigure}{0.49\columnwidth}
    \includegraphics[width=\columnwidth]{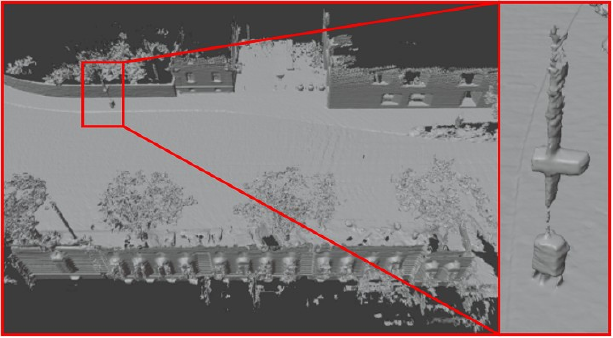}
    \subcaption{Voxel size: $5$ cm}
    \label{fig:reconstruction_our_1}
    \end{subfigure}
    \begin{subfigure}{0.49\columnwidth}
    \includegraphics[width=\columnwidth]{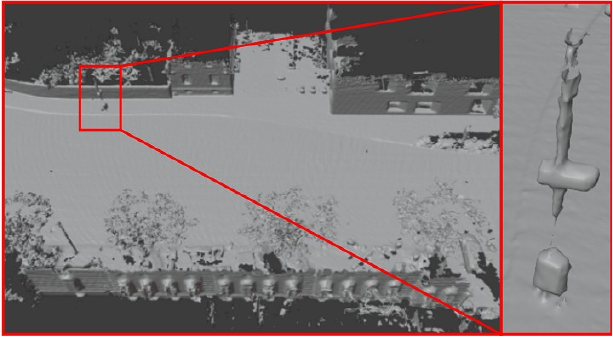}
    \subcaption{Voxel size: $10$ cm}
    \label{fig:reconstruction_our_2}
    \end{subfigure}
    \begin{subfigure}{0.49\columnwidth}
    \includegraphics[width=\columnwidth]{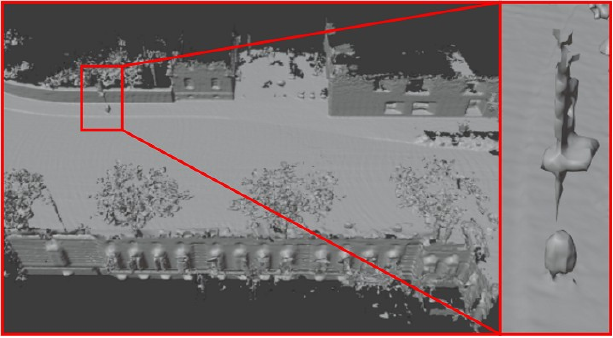}
    \subcaption{Voxel size: $15$ cm}
    \label{fig:reconstruction_our_3}
    \end{subfigure}
    \begin{subfigure}{0.49\columnwidth}
    \includegraphics[width=\columnwidth]{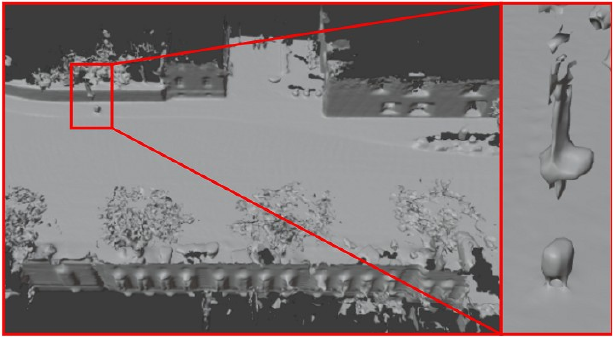}
    \subcaption{Voxel size: $20$ cm}
    \label{fig:reconstruction_our_4}
    \end{subfigure}
    \caption{
    3D reconstruction on our dataset with different voxel size and an adaptive truncation distance $\varepsilon_\mathrm{apt}:[40, 80]$ cm.
    }
    \label{fig:reconstruction_ours}
\end{figure}

In LGSVL \cite{rong2020lgsvl}, we simulate an experiment vehicle moving around a SUV at a distance of $10$ m and a velocity of $5$ m/s to evaluate the reconstruction accuracy of generated models.
The experiment vehicle is equipped with a top-mounted 128-beam LiDAR $2.3$ m above ground.
The rays are specialized according to the Velodyne VLS-128 calibration with a frequency of $10$ Hz.
Additionally, we add white noise to the simulated ranges with a standard deviation of $1$ cm.
In this experiment, we use a voxel size of $2.5$ cm and a truncation distance of $\varepsilon=10$ cm, $\varepsilon=30$ cm or an adaptive truncation distance range $\varepsilon_\mathrm{apt}:[10, 30]$ cm.
The results in Fig. \ref{fig:reconstruction_suv} shows that a large $\varepsilon$ ensures the reconstructed surface's completeness, while a small $\varepsilon$ retains much model details.
Fig. \ref{fig:reconstruction_suv_comp} illustrates the quantitative accuracy evaluation of the reconstructed models.
It shows that the reconstructed surface using adaptive TSDF is smoother and more accurate than the surface generated by standard TSDF.
In Fig. \ref{fig:suv_a_comp}, the maximum distance error between the reconstructed surfaces and ground truth is $11$ cm, while in Fig. \ref{fig:suv_b_comp} only $6$ cm.
Besides that, we also evaluate our adaptive TSDF-based 3D reconstruction on the real world KITTI dataset.
As shown in Fig. \ref{fig:reconstruction_kitti}, by benefiting from our adaptive data fusion strategy the reconstructed model in Fig. \ref{fig:kitti_atsdf} has a higher resolution and contains more details.
Furthermore, to evaluate the effect of voxel size on the reconstruction's quality, we design four experiments with a voxel size of $5$ cm, $10$ cm, $15$ cm, and $20$ cm using our adaptive TSDF-based algorithm on the dataset recorded with our experiment vehicle.
The truncation distance $\varepsilon_\mathrm{apt}$ is dynamically computed in the range of $[40, 80]$ cm.
The results in Fig. \ref{fig:reconstruction_ours} show that a smaller voxel size can not significantly improve the quality of the reconstructed models because of the measurement errors of the LiDAR sensor and the slightly inaccurate sensor poses.
However, a too large voxel size is not friendly to smaller objects, as illustrated in Fig. \ref{fig:reconstruction_our_3} and Fig. \ref{fig:reconstruction_our_4}.
Since a small voxel size means a large amount of faces, it is necessary to choose a suitable voxel size according to different tasks and sensor specifics.
In the rest of this chapter, if without special declarations, the voxel size is set as $10$ cm and $\varepsilon_\mathrm{apt}$ is dynamically computed in the range of $[40, 80]$ cm.
\begin{figure}
    \vspace{0.2cm}
    \centering
    \begin{subfigure}{0.49\columnwidth}
    \includegraphics[width=\columnwidth]{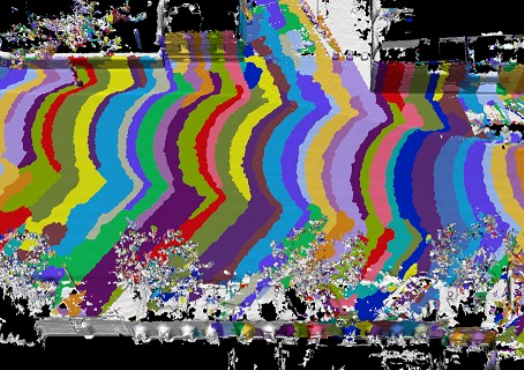}
    \subcaption{Mesh texturing}
    \label{fig:eva_view_selection}
    \end{subfigure}
    \begin{subfigure}{0.49\columnwidth}
    \includegraphics[width=\columnwidth]{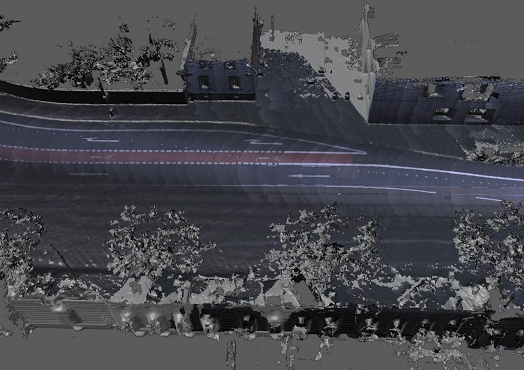}
    \subcaption{Without color adjustment}
    \label{fig:texture_vignetting}
    \end{subfigure}
    \begin{subfigure}{0.49\columnwidth}
    \includegraphics[width=\columnwidth]{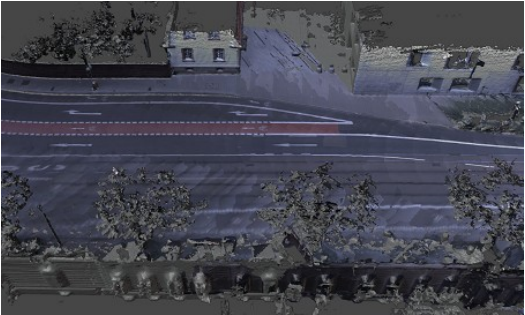}
    \subcaption{Vignetting coorection}
    \label{fig:texture_devignetting}
    \end{subfigure}
    \begin{subfigure}{0.49\columnwidth}
    \includegraphics[width=\columnwidth]{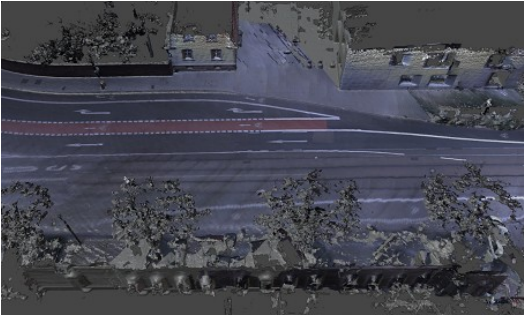}
    \subcaption{With color adjustment}
    \label{fig:texture_withall}
    \end{subfigure}
    \caption{
    Model texturing and color optimization.
    }
    \label{fig:texture_map}
\end{figure}
\begin{figure}
    \centering
    \begin{subfigure}{0.49\columnwidth}
    \includegraphics[width=\columnwidth]{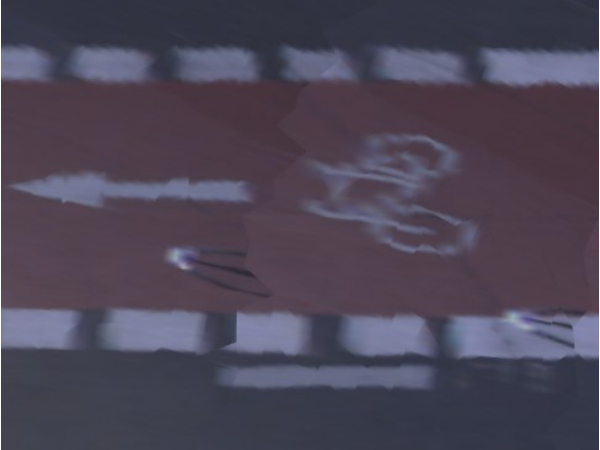}
    \subcaption{}
    \end{subfigure}
    \begin{subfigure}{0.49\columnwidth}
    \includegraphics[width=\columnwidth]{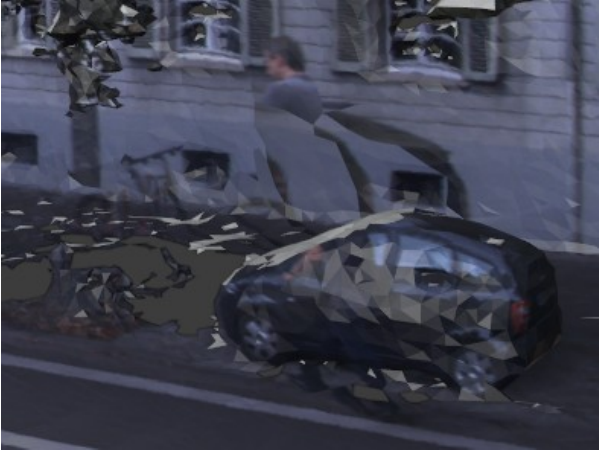}
    \subcaption{}
    \end{subfigure}
    \begin{subfigure}{0.49\columnwidth}
    \includegraphics[width=\columnwidth]{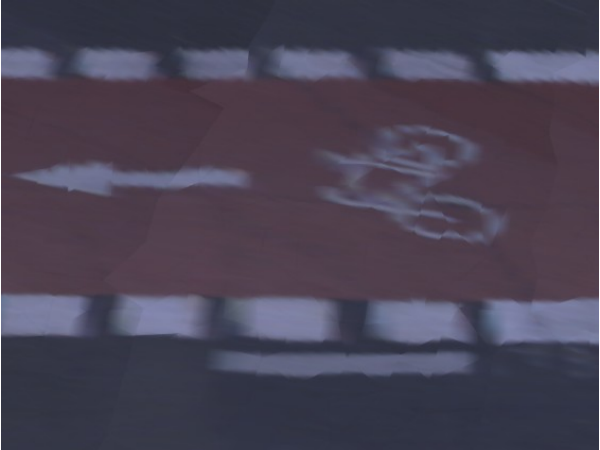}
    \subcaption{}
    \end{subfigure}
    \begin{subfigure}{0.49\columnwidth}
    \includegraphics[width=\columnwidth]{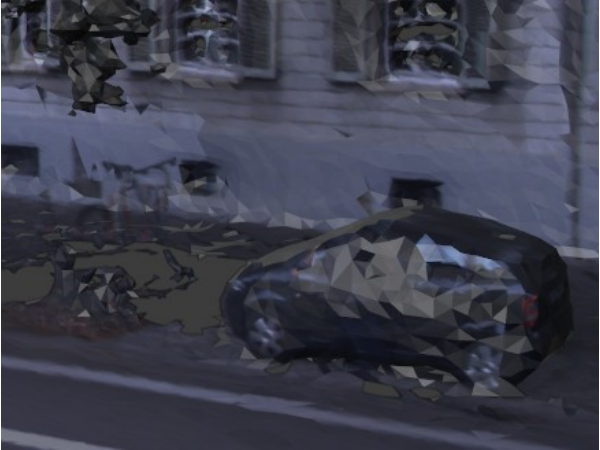}
    \subcaption{}
    \end{subfigure}
    \caption{
    The situations presented in (a) and (b) illustrate the wrong textures caused by occlusions.
    (c) and (d) are the corrected textures after the photo consistency check.
    }
    \label{fig:texture_photocheck}
\end{figure}
\subsection{Texture Mapping}
To evaluate the texture mapping process, we use the dataset recorded with our experimental vehicle.
As argued previously, textures obtained from the view selection contain strong color discontinuities and wrong texture patches caused by occlusions, as shown in Fig. \ref{fig:texture_vignetting}.
To handle the vignetting issue of image pixels, we perform vignetting correction for images as a preprocessing step.
Fig. \ref{fig:texture_devignetting} shows that the vignetting correction algorithm can significantly improve the texture's quality.
Further, the effect of the photo consistency check is shown in Fig. \ref{fig:texture_photocheck}.
In Fig. \ref{fig:texture_withall} we show the texturing result with the global color adjustment technique.

\subsection{Semantic Mapping}
\begin{figure}
    \centering
    \begin{subfigure}{0.32\columnwidth}
    \includegraphics[width=\columnwidth]{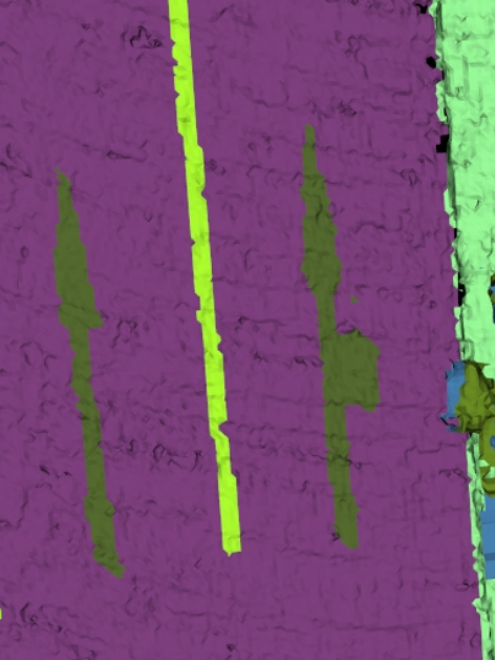}
    \subcaption{Voxel size:$5$cm.}
    \end{subfigure}
    \begin{subfigure}{0.32\columnwidth}
    \includegraphics[width=\columnwidth]{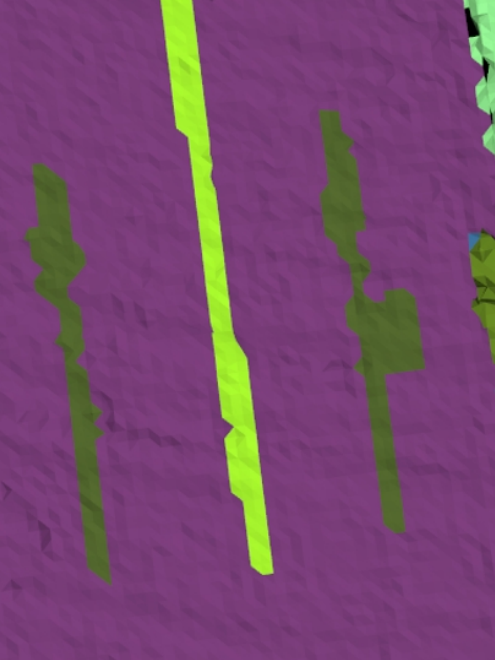}
    \subcaption{Voxel size:$10$cm.}
    \end{subfigure}
    \begin{subfigure}{0.32\columnwidth}
    \includegraphics[width=\columnwidth]{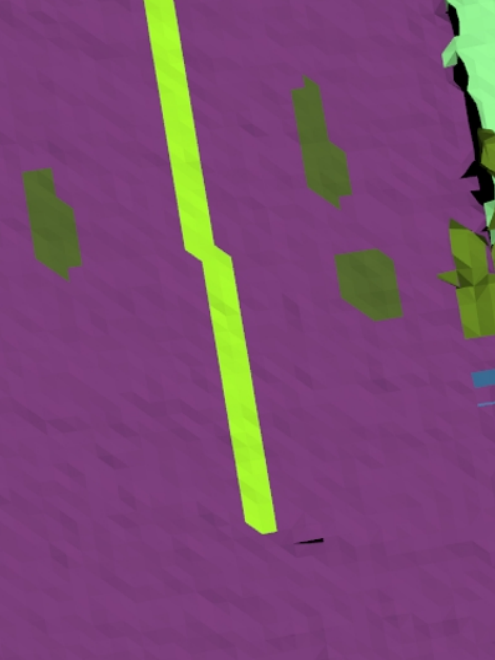}
    \subcaption{Voxel size:$15$cm.}
    \end{subfigure}
    \caption{Road marking with different voxel sizes.}
    \label{fig:semantic_voxel_size}
\end{figure}

\begin{figure}
    \centering
    \begin{subfigure}{\columnwidth}
    \includegraphics[width=\columnwidth]{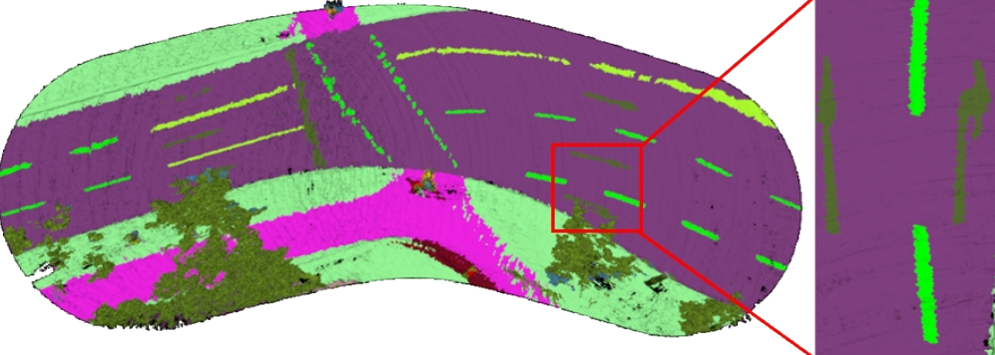}
    \subcaption{Data term only.}
    \end{subfigure}
    \begin{subfigure}{\columnwidth}
    \includegraphics[width=\columnwidth]{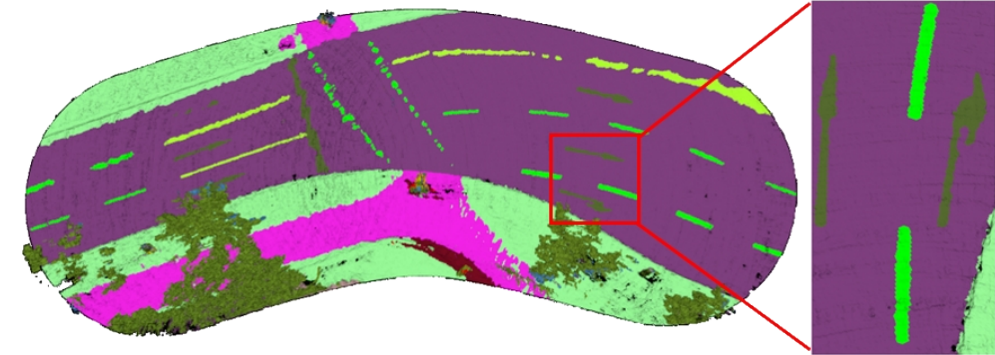}
    \subcaption{Data term + smoothing term.}
    \end{subfigure}
    \caption{MRF in comparison to naive approach.}
    \label{fig:MRF}
\end{figure}

\begin{figure}
    \centering
    \begin{subfigure}{\columnwidth}
    \includegraphics[width=\columnwidth]{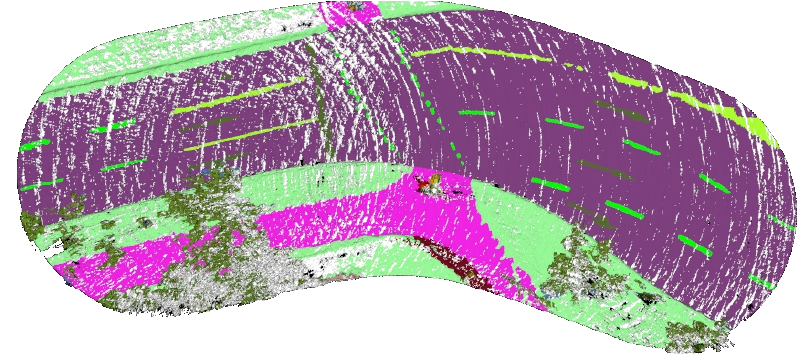}
    \subcaption{Original mesh map with white faces.}
    \end{subfigure}
    \begin{subfigure}{\columnwidth}
    \includegraphics[width=\columnwidth]{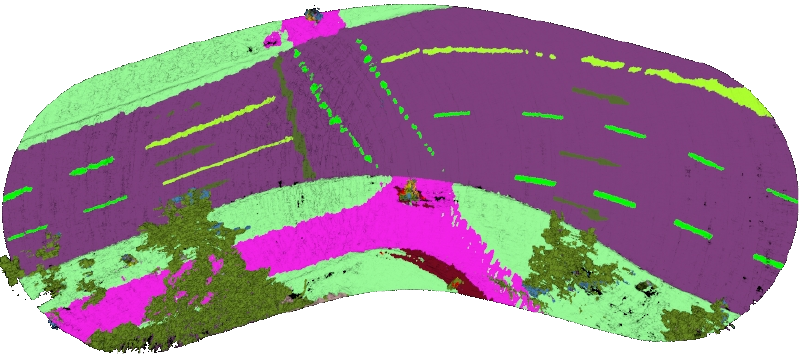}
    \subcaption{Mesh map processed with MRF.}
    \end{subfigure}
    \caption{Elimination of invisible white faces with MRF.}
    \label{fig:padding}
\end{figure}

Basically, in order to create accurate semantic mapped models, the voxel size for the 3D geometric reconstruction should be as fine as possible.
Road markings such as dashed lines and arrows need extremely fine voxel size to be reconstructed.
As shown in the Fig. \ref{fig:semantic_voxel_size}, a voxel size of $5$ cm is a good choice although it results a large number of faces.
Using bigger voxel sizes, it is worse to recognize the arrow markings with a reasonable accuracy.
Secondly, we evaluate the MRF-based fusion approach compared to a naive method, which only consider the data term and chose the label with the highest probability.
As illustrated in Fig. \ref{fig:MRF}, the edge of the road markings are reconstructed smoother using the MRF-based method with less outliers.
Another advantage of the MRF-based approach is to estimate the semantic class for the invisible faces.
As shown in Fig. \ref{fig:padding}(b), all the semantic class of the invisible faces are greatly estimated with the help of the smoothing term $E_\mathrm{S}$ presented in Eq. \ref{equ:semantic_mrf_energy_function}.

%% file: 08_conclusions.tex
\section{CONCLUSION AND FUTURE WORK}
\label{sec:CONCLUSION AND FUTURE WORK}
In this work, we present a large-scale 3D reconstruction, texturing and semantic mapping system.
For the 3D reconstruction, an adaptive TSDF is proposed to tackle the heterogeneous density of LiDAR point clouds.
The texture mapping procedure using an optimal image patch selection and color optimization techniques has a high efficiency and robustness.
Using a MRF-based method, geometric models can be accurate and seamless semantic textured.  
By applying our approach, several applications for autonomous driving are enabled.
The accurately textured 3D models can be used as a realistic simulation environment.
Compared to synthetic simulators, our models are more similar to the real world and can accelerate the algorithm developing and testing sufficiently.
Besides that, the semantic mapped models can be used to generate training data automatically for deep learning algorithms.
In addition, it is also very helpful for the automatic generation of semantic HD maps for automated driving vehicles.
As future works we focus on:
Adapting the voxel size for the reconstruction to different semantic classes.
Extracting semantic HD maps from the semantic models automatically for autonomous driving.
Generating training data for deep learning approaches using the semantic models.

%% file: 09_acknowledgements.tex
\section*{Acknowledgment}
This research is accomplished within the project “UNICARagil”.
We acknowledge the financial support for the project by the Federal Ministry of Education and Research of Germany (BMBF).